# APPLYING PHONOLOGICAL FEATURES IN MULTILINGUAL TEXT-TO-SPEECH

*Cong Zhang[1]*   *Huinan Zeng[2]*   *Huang Liu[3]*   *Jiewen Zheng[3]*

[1] Centre for Language Studies, Radboud University, Nijmegen, Netherlands;
[2] Faculty of Linguistics, Philology and Phonetics, University of Oxford, Oxford, UK;
[3] Shengrurenxin Ltd., Beijing, China

## ABSTRACT

This study investigates whether phonological features can be applied in text-to-speech systems to generate native and non-native speech in English and Mandarin. We present a mapping of *ARPABET/pinyin* to *SAMPA/SAMPA-SC* and then to *phonological features.* We tested whether this mapping could lead to the successful generation of native, non-native, and code-switched speech in the two languages. We ran two experiments, one with a small dataset and one with a larger dataset. The results proved that phonological features could be used as a feasible input system, although further investigation is needed to improve model performance. The accented output generated by the TTS models also helps with understanding human second language acquisition processes.

## 1. INTRODUCTION

Many latest modelling experiments in text-to-speech (TTS) or commercial TTS systems adopt language-specific phone labels such as ARPABET (a.k.a. CMU phones) for English, and pinyin for Mandarin Chinese. However, while they can achieve high performance, each individual language requires a different phone set. When training models for any new or additional language, the front-end system often has to alter extensively to suit the input of the new or additional language. The changes bring extra workload and delays the development progress whenever a new language is added. To optimise the workflow, a more universal input representation is needed for as many languages as possible. Therefore, we propose an input system based on phonological feature theories with an aim to progress towards a universal TTS system. This is especially beneficial for under-resourced languages which have limited amount of available data or well-defined input labels.

Previous works have investigated the possibility of applying phonological features in TTS. [1] synthesised in a language that was not present in the input data. While their results were encouraging, their investigation was mainly about vowels. [2] examined a more complex set of features and applied it to more languages with small amount of data. The current study differentiates from past studies in that it is based on a well-examined theoretically driven phonological model, *Featurally Underspecified Lexicon* (FUL) [3], following its successful application in an automatic speech recognition (ASR) system [4]. FUL has also been tested extensively in human speech perception [5] and production [6], as well as child [7] and L2 language acquisition [8].

Past studies such as [1], [9] have examined typologically similar languages, while the current study investigates two typologically distant languages, English and Mandarin. Besides applying features to the segment level, this study also presents an extra dimension of difficulty with the lexical tone specification in Mandarin and the lack of it in English.

The benefits of using phonological features rather than language-specific labels, or the more universal IPA (International Phonetic Alphabet) are that: (1) a much smaller and limited number of features are used to represent all languages. Even though IPA is already a smaller set of labels than merging language-specific labels from different languages, it is still expandable based on the languages and would results in more than 60 labels. Phonological features amount to a much smaller number. For instance, the current study proposes a set of 19 phonological features in total. (2) A unified feature set for multiple languages means being able to pool data from different languages together and therefore reducing the amount of data needed for model training, as well as solving the data sparsity issue. (3) It fuses multiple languages across speakers into a unified acoustic model, making the deployment for production much easier. One specific instance is that when generating code-switch speech, the timbre would be the same in different languages.

In this paper, we present a mapping between the widely adopted ARPABET (English) and pinyin (Mandarin) input, and their mapping with phonological features through the intermediate coding of SAMPA [10] for English and SAMPA-SC [11] for Mandarin. The mapping from *ARPABET/pinyin* to *SAMPA/SAMPA-SC* and then to *phonological features* has been made publicly available in [12]. This direct mapping makes the results from this study directly applicable to existing TTS systems which take ARPABET and pinyin labels as input and already have a substantial amount of labelled data. Moreover, apart from aiming to test the feasibility of applying phonological features in TTS, we also have a theoretical goal of using the findings of this study to shed light on human phonological acquisition process.

## 2. PHONOLOGICAL FEATURES

### 2.1. Phonological feature theories

Distinctive features are designed to describe the phonological systems in the world's languages. The idea was first proposed by [13]. Since then, it has received continued attention among phonologists and different feature organisations have been proposed. See [14] for a review of different feature theories.

The feature geometry in FUL is shown in Figure 1. There is only a handful of features in this system. The manner of articulation is represented by the ROOT features, CONSTRICTION features, and [STRIDENT, NASAL, LATERAL, RHOTIC]. LARYNGEAL features mark the voice of articulation and the place of articulation is indicated by the PLACE features. The FUL model is used for the following reasons:

1. All features are monovalent and there are no dependent features. Comparing to systems where there are binary features and feature dependencies, FUL's feature marking is more straightforward. Monovalent keeps the dimensionality of the final input vector to a minimum.

2. Consonants and vowels share PLACE features. No extra feature is needed to describe vowels. For example, labial consonants and rounded vowels are [LABIAL], coronal consonants and front vowels are [CORONAL], and dorsal consonants and back vowels are [DORSAL]. For consonants, the TONGUE HEIGHT and TONGUE ROOT features become relevant when there are contrasts within the same ARTICULATOR. For instance, contrasts within coronal sounds e.g. dentals, palatals vs. retroflexes, can be established by a combination of [CORONAL] and TONGUE HEIGHT features. The number of features is therefore also kept to a minimum.

3. The features in the geometry have acoustic correlates. For example, the TONGUE HEIGHT feature [HIGH] is characterised by the concentration of more energy at higher frequencies and low F1 for vowels, and the feature [LOW] is characterised by a concentration of more energy at lower frequencies, and high F1 for vowels.

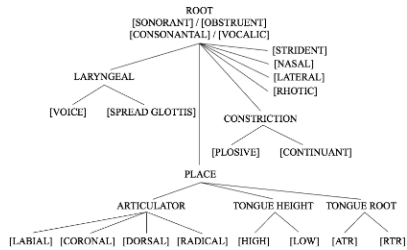

**Figure 1**. Feature organisation in the FUL model [15]

### 2.2. Feature sets for Mandarin and English
*2.2.1. English*
The features for the phonemes in General American English in this study are shown in Table 1. The parentheses indicate that the feature specification for the phone is optional as it is not required to establish contrast within the language, but it is specified in this study since our aim is to build a cross-linguistic system which can distinguish both English and Mandarin. There are 38 phonemes, within which 24 are [CONSONANTAL] and 14 are [VOCALIC]. Consonants are further divided up by the ROOT features [OBSTRUENT] and [SONORANT] into obstruents /b p v f d t dʒ tʃ ð θ s z ʃ ʒ g k h/ and sonorants /m n ŋ w j l ɹ/ respectively.

Obstruents contrast in voicing. Voiced obstruents /b v d dʒ ð z ʒ g/ are specified for [VOICE] and voiceless obstruents /p f t tʃ θ s ʃ k h/ are not specified for LARYNGEAL features. The plosives, fricatives, and affricates are distinguished by the feature [STRIDENT] and the CONSTRICTION features [PLOSIVE] and [CONTINUANT]. Plosives /b p d t g k/ are [PLOSIVE]. Fricatives /v f ð θ s z ʃ ʒ h/ are [CONTINUANT], with /s z ʃ ʒ/ also being [STRIDENT]. Affricates /dʒ tʃ/ are [PLOSIVE, STRIDENT]. All sonorant consonants are specified for [VOICE]. /m n ŋ/ are [NASAL], /l/ is [LATERAl] and /ɹ/ is [RHOTIC]. The main place of articulation for the consonants is represented by the ARTICULATOR features under the PLACE node. The coronal fricative phonemes /s z ʃ ʒ/ contrast in their specification for TONGUE HEIGHT features – /ʃ ʒ/are [HIGH] and /s z/ are not specified for TONGUE HEIGHT. Vowels share the features under the PLACE node with consonants. The front vowels /i ɪ e ɛ æ/ are [CORONAL], rounded back vowels /u ʊ ɔ o/ are [LABIAL, DORSAL] and the unround back vowel /ɑ/ is [DORSAL]. The high vowels /u ʊ i ɪ/ are [HIGH], low vowels /æ ɑ/ are [LOW], and mid vowels /ɔ o e ɛ ə ɚ ɝ ʌ/ are not specified for TONGUE HEIGHT. General American English also makes use of the TONGUE ROOT features to establish the tense /u o i e ʌ/ [ATR] and lax vowel /ʊ ɔ ɪ ɛ ɑ ə æ/ [RTR] contrast. The vowels /e o/ are often realised as the diphthongs [eɪ oʊ] respectively but they are usually considered as tensed monophthong phonemes. The true diphthongs in General American English are /aɪ aʊ ɔɪ/. There are two [RHOTIC] vowels /ɚ ɝ/.

**Table 1.** Features for English phonemes

| Features | English phonemes |
|---|---|
| CONSONANTAL | b p v f d t dʒ tʃ ð θ s z ʃ ʒ g k h m w n j l ɹ ŋ |
| VOCALIC | u ʊ ɔ o i ɪ e ɛ æ ɑ ə ɚ ɝ ʌ |
| SONORANT | m w n j l ɹ ŋ u ʊ ɔ o i ɪ e ɛ æ ɑ ə ɚ ɝ ʌ |
| OBSTRUENT | b p v f d t dʒ tʃ ð θ s z ʃ ʒ g k h |
| VOICE | b v d dʒ ð z ʒ g m w n j l ɹ ŋ u ʊ ɔ o i ɪ e ɛ æ ɑ ə ɚ ɝ ʌ |
| PLOSIVE | b p d t dʒ tʃ g k |
| CONTINUANT | v f ð θ s z ʃ ʒ h |
| LABIAL | b p v f m w u ʊ ɔ o |
| CORONAL | d t dʒ tʃ ð θ s z ʃ ʒ n j l ɹ i ɪ e ɛ æ |
| DORSAL | g k w ŋ u ʊ ɔ o ɑ |
| RADICAL | h |
| HIGH | ʃ ʒ u ʊ i ɪ (dʒ tʃ g k w j) |
| LOW | æ ɑ |
| ATR | (u o i e ʌ) |
| RTR | ʊ ɔ ɪ ɛ ɑ ə (æ) |
| NASAL | m n ŋ |
| LATERAL | L |
| STRIDENT | dʒ tʃ s z ʃ ʒ |
| RHOTIC | ɹ ɚ ɝ |

*2.2.2. Mandarin*

Table 2 shows the features for the 37 Mandarin phonemes. There are 21 phonemes that are [CONSONANTAL] and 16 phonemes that are [VOCALIC]. The overall number of sounds as well as the number of vowels and consonants are similar to that of General American English.

Different from English, Mandarin obstruents contrast in aspiration instead of voicing. The aspirated obstruents /pʰ tʰ tsʰ tʂʰ kʰ/ are specified for [SPREAD GLOTTIS] and the unaspirated ones [p f t s ɕ ts tʂ k x] are not specified for LARYNGEAL features. The only voiced obstruent /z̩/ is analysed as a syllabic approximant ([16]) or an apical vowel ([17]) in other studies. This study follows [16] and treat it as /z̩/ [CONSONANTAL, OBSTRUENT, VOICE]. Mandarin has a three-way coronal contrast – dental [s ts tsʰ] vs. alveolo-palatal [ɕ tɕ tɕʰ] vs. retroflex [ʂ tʂ tʂʰ]. The alveolo-palatals and dentals are in complementary distribution. The dentals are the phonemes /s ts tsʰ/ and the alveolo-palatals are the surface allophones [ɕ tɕ tɕʰ]. The phonemes realise as [ɕ tɕ tɕʰ] when they occur before the high front vowels /i y/ and as [s ts tsʰ] in other phonological environments. The dental and retroflex phonemes are distinguished by the TONGUE HEIGHT features. The retroflexes are marked as [RTR] and the dentals are not specified for TONGUE HEIGHT. The Mandarin sonorant consonants /m n ŋ w j l/ have the same features as those in English. Mandarin does not have [RHOTIC] consonant.

Mandarin has a larger number of nasalised vowels and rhotic vowels than English. /u˞ o˞ ɛ˞ a˞ ɚ/ are [RHOTIC] and /ã˞ ɚ̃ ũ˞/ are [NASAL, RHOTIC]. Besides the back rounded vowels that also exist in English, Mandarin has a front rounded vowel /y/ [LABIAL, CORONAL].

**Table 2.** Features for Mandarin phonemes

| Features | Mandarin phonemes |
|---|---|
| CONSONANTAL | p pʰ f t tʰ s ʂ z̩ ts tsʰ tʂ tʂʰ k kʰ x m w n j l ŋ |
| VOCALIC | i y u u˞ ũ˞ o o˞ e ɛ ɛ˞ a a˞ ã˞ ə ɚ ɚ̃ |
| SONORANT | m w n j l ŋ i y u u˞ ũ˞ o o˞ e ɛ ɛ˞ a a˞ ã˞ ə ɚ ɚ̃ |
| OBSTRUENT | p pʰ f t tʰ s ʂ z̩ ts tsʰ tʂ tʂʰ k kʰ x |
| VOICE | z̩ m w n j l ŋ i y u u˞ ũ˞ o o˞ e ɛ ɛ˞ a a˞ ã˞ ə ɚ ɚ̃ |
| SPREAD GLOTTIS | pʰ tʰ tsʰ tʂʰ kʰ |
| PLOSIVE | p pʰ t tʰ ts tsʰ tʂ tʂʰ k kʰ |
| CONTINUANT | f s ʂ z̩ x |
| LABIAL | p pʰ f m w y u u˞ ũ˞ o o˞ |
| CORONAL | t tʰ s ʂ z̩ ts tsʰ tʂ tʂʰ n j l i y e ɛ ɛ˞ |
| DORSAL | k kʰ x w ŋ u u˞ ũ˞ o o˞ |
| HIGH | ʂ z̩ tʂ tʂʰ i y u u˞ ũ˞ (k kʰ w j) |
| LOW | a a˞ ã˞ |
| ATR | (i y u u˞ ũ˞ o o˞ e) |
| RTR | ʂ z̩ tʂ tʂʰ ɛ ɛ˞ ə ɚ ɚ̃ |
| NASAL | m n ŋ ũ˞ ã˞ ɚ̃ |
| LATERAL | l |
| STRIDENT | s ʂ z̩ ts tsʰ tʂ tʂʰ |
| RHOTIC | u˞ ũ˞ o˞ ɛ˞ a˞ ã˞ ɚ ɚ̃ |

## 3. EXPERIMENTS

In this study, we followed a modified model structure based on FastSpeech [18]. We replaced the phoneme input in the original model with text input; and then used the FastSpeech acoustic model to generate mel-spectrogram from text. The input text contained SAMPA phonemes [10], prosody features, and tone feature. The SAMPA phonemes were then converted into corresponding binary feature using the mapping in [12]. The prosody and tone features were also converted to prosody and tone embeddings respectively. The binary feature and the two embeddings were then concatenated as a text embedding, and was used as the input for the FastSpeech feed-forward transformer. The temporal alignment for the phonmes were obtained through Kaldi [19] ASR tool.

Our modified FastSpeech model consisted of 4 FFT blocks on both the phoneme side and the mel-spectrogram side. For the text, the SAMPA phoneme vocabulary size was X1(including punctuations); the size of the prosody feature was 4; and the size of the tone feature was 6. The phonemic binary feature was used as the input to a dense layer, and the output size was set to 192. The output of prosody embedding and tone embedding sizes were set to 32 and 32 respectively. The final linear layer output a 128-dimensional mel-spectrogram. All the other parameters were set following what were reported in [18]. We trained our modified FastSpeech model on a NVIDIA GeForce RTX 2080 GPU. We used batch size 24 for training, and it took 200k steps for training until convergence. The optimizer and other hyper-parameters were the same as those reported in [18]. The output mel-spectrograms of our FastSpeech model were transformed into audio samples using a pretrained MelGan vocoder [20]. Demos for the following two experiments are available at:
https://congzhang365.github.io/feature_tts/.

### 3.1 Experiment 1: small dataset

In this experiment, we tested whether using phonological features as input could generate intelligible speech with small amount of data. We examined both the generation of the input language and an unseen language.

*3.1.1. Dataset*

The data used in this study were studio-recorded high-quality data, with a sampling rate of 24k Hz, at 16-bit. A set of Mandarin data and a set of English data were used. The English data were recorded by a male, native American English speaker (EN_MS1) in the style of AI assistant, with a total length of 2.62 hours. The Mandarin data were recorded by a male, native Mandarin speaker (CN_MS1) in the style of news reading, with a total length of 20 hours. To test the performance with small data, we designed three intervals of data input:

- M0.5: 2.62 hours of English, and 0.5 hours of Mandarin
- M2: 2.62 hours of English, and 2 hours of Mandarin
- M8: 2.62 hours of English, and 8 hours of Mandarin

Two input speakers were chosen to generate 34 Mandarin-only, English-only, and Mandarin-English code-mixed test utterances. Since our investigation was still at an early stage, we chose to evaluate the results by making linguistic analysis of the output audio. An experienced phonetician and TTS linguist analysed the output and reported the results and analyses as objectively as they could.

*3.1.2. Results and analyses*

*English utterances*: Since the input data contained 2.62h of native English speech, the English speaker, EN_MS1, could produce the English test utterances well from `M0.5` onwards. However, the Mandarin speaker, CN_MS1, while having English-like rhythm, could not produce intelligible English utterances in any model.

*Mandarin utterances*: In `M0.5`, CN_MS1 could only vaguely produce the Mandarin utterances. Some words were intelligible, but not all. Both `M0.5` and `M2` presented issues with voice quality, lexical tone accuracy, and intonation. In `M8`, CN_MS1 could reliably produce intelligible utterances, despite some small voice quality issues. EN_MS1 could not produce intelligible Mandarin utterances in any of the three models. However, if a listener knew the text of the utterance beforehand, the words could be identified very well.

*Mandarin-English code-mixed utterances:* In `M0.5`, both EN_MS1 and CN_MS1 produced identifiable words, with the English speaker being able to produce clearer utterances. `M2` presented improvement over `M0.5`. In `M8`, both speakers produced much clearer utterances in code-switching mode. EN_MS1's output was near ideal, except for lacking lexical tone specification on the Mandarin words.

**3.2 Experiment 2: larger multi-speaker dataset**

In this experiment, we used a larger dataset with more speakers and data in both English and Mandarin, in order to test whether using phonological features as input can generate more intelligible output in Mandarin, English, as well as code-mixed utterances with both language.

*3.2.1. Dataset*

A 100-hour multi-speaker dataset was used in this experiment. The dataset contained speech from twelve speakers, including nine native Mandarin speakers and three native American English speakers. The native English speakers' data totalled 9 hours of native General American speech. The Mandarin speakers' data included 5 hours of second-language English speech, 10 hours of code-mixing speech between Mandarin and English, and 76 hours of native Mandarin speech. The test utterances and evaluation procedures in Experiment 1 were also used in this experiment. The following speakers and their data were chosen to generate the test sentences.

- CN_FS1: Mandarin-English cross-lingual data by a female native Mandarin speaker. The data consisted of 15 hours of speech in total, including 10 hours of Mandarin speech, 2 hours of English speech, and 3 hours of Mandarin-English code-switching speech. The style and speech materials were designed to build an AI-assistant.
- CN_MS2: 9 hours of native Mandarin data by a male native Mandarin speaker. The speech material and style were designed to make announcements and advertise.
- EN_MS1: 2.62 hours of English data by a male native American English speaker. The style and speech materials were designed to build an AI-assistant.

*3.2.2. Results and analyses*

Trained with a larger corpus with multi-lingual and multi-speaker data, the outputs from this experiment was much more ideal. All utterances were intelligible and the styles and sentence prosodies were successfully acquired. CN_MS2, which did not have English data or code-switching data in the input, performed less well than the other two speakers on both the English utterances and the code-switching utterances. EN_MS1's output still missed the majority of the lexical tones, although it became much more fluent and identifiable than in the previous experiment. CN_FS1 produced good English utterances and code-switching utterances. However, it is interesting to note that the English outputs from both Mandarin speakers carried heavy Chinese accent.

## 4. DISCUSSION AND CONCLUSIONS

The results showed that using phonological features as input in TTS was indeed feasible, although the results were not yet ideal. One potential reason for the suboptimal performance was at the forced alignment stage. Since the forced alignment tool was trained on an ARPABET or pinyin level, while the text input was based on a more fine-grained phoneme set, SAMPA, errors might have occurred in the forced alignment process, leading to bad cases that made the output less intelligible. We thus are also developing more accurate ways of automatically aligning the audios to phones [21].

Apart from the possibility of using phonological features in TTS application, the findings of this study also provided supporting evidence for phonological theories. In distinctive feature theory, listeners first extract the features from the auditory input data. The extracted features are then mapped onto the underlying contrasts of the listeners' native language. The listeners/speakers then add the phonological rules and processes of their native language to produce non-native speech [8]. The accentedness in second language production can be explained by this theory. Our findings of English speaker lacking tonal contrasts and Mandarin speaker having Mandarin-like vowels in English utterances both support this process. Typologically, the investigation of tonal and non-tonal languages in this study contributed to the general investigation of applying phonological features in TTS (c.f. [1], [9]) for languages with different ways of prosodic specification on a lexical level.